%% file: VideoTitling.tex
\documentclass[sigconf]{acmart}

\AtBeginDocument{%
  \providecommand\BibTeX{{%
    \normalfont B\kern-0.5em{\scshape i\kern-0.25em b}\kern-0.8em\TeX}}}

\copyrightyear{2020}
\acmYear{2020}
\setcopyright{acmcopyright}\acmConference[KDD '20]{Proceedings of the 26th ACM SIGKDD Conference on Knowledge Discovery and Data Mining}{August 23--27, 2020}{Virtual Event, CA, USA}
\acmBooktitle{Proceedings of the 26th ACM SIGKDD Conference on Knowledge Discovery and Data Mining (KDD '20), August 23--27, 2020, Virtual Event, CA, USA}
\acmPrice{15.00}
\acmDOI{10.1145/3394486.3403325}
\acmISBN{978-1-4503-7998-4/20/08}

\usepackage{multirow}

\begin{document}

\title{Comprehensive Information Integration Modeling Framework for Video Titling}

\author[S. Zhang*, Z. Tan*, J. Yu, Z. Zhao, K. Kuang, T. Jiang, J. Zhou, H. Yang, F. Wu]{
    Shengyu Zhang$^{1*}$, Ziqi Tan$^{1*}$, Jin Yu$^{2}$, Zhou Zhao$^{1\dagger}$, Kun Kuang$^{1}$, Tan Jiang$^{1}$, Jingren Zhou$^{2}$, Hongxia Yang$^{2\dagger}$, Fei Wu$^{1\dagger}$
}
\affiliation{
    $^1$ College of Computer Science and Technology, Zhejiang University
}
\affiliation{
    $^2$ Alibaba Group
}
\email{
  {sy_zhang, tanziqi, zhaozhou, kunkuang, jiangtan, wufei}@zju.edu.cn
}
\email{
  {kola.yu, jingren.zhou, yang.yhx}@alibaba-inc.com
}

\renewcommand{\authors}{Shengyu Zhang, Ziqi Tan, Jin Yu, Zhou Zhao, Kun Kuang, Tan Jiang,  Jingren Zhou, Hongxia Yang, Fei Wu}

\renewcommand{\shortauthors}{S. Zhang*, Z. Tan*, J. Yu, Z. Zhao, K. Kuang, T. Jiang, J. Zhou, H. Yang, F. Wu}
\newcommand{\model}{ model }
\newcommand{\etal}{\textit{et al}.}
\newcommand{\ie}{\textit{i}.\textit{e}.}
\newcommand{\eg}{\textit{e}.\textit{g}.}
\newcommand{\Our}{Gavotte   }
\newcommand{\vpara}[1]{\vspace{0.05in}\noindent\textbf{#1 }}

\begin{abstract}
In e-commerce, consumer-generated videos, which in general deliver consumers' individual preferences for the different aspects of certain products, are massive in volume. To recommend these videos to potential consumers more effectively, diverse and catchy video titles are critical. However, consumer-generated videos seldom accompany appropriate titles. To bridge this gap, we integrate comprehensive sources of information, including the content of consumer-generated videos, the narrative comment sentences supplied by consumers, and the product attributes, in an end-to-end modeling framework. Although automatic \textit{video titling} is very useful and demanding, it is much less addressed than video captioning. The latter focuses on generating sentences that describe videos as a whole while our task requires the product-aware multi-grained video analysis. To tackle this issue, the proposed method consists of two processes, 
\ie, granular-level interaction modeling and abstraction-level story-line summarization. Specifically, the granular-level interaction modeling first utilizes temporal-spatial landmark cues, descriptive words, and abstractive attributes to builds three individual graphs and recognizes the intra-actions in each graph through Graph Neural Networks (GNN). Then the global-local aggregation module is proposed to model inter-actions across graphs and aggregate heterogeneous graphs into a holistic graph representation. The abstraction-level story-line summarization further considers both frame-level video features and the holistic graph to utilize the interactions between products and backgrounds, and generate the story-line topic of the video. We collect a large-scale dataset accordingly from real-world data in Taobao, a world-leading e-commerce platform, and will make the desensitized version publicly available to nourish further development of the research community\footnote{Dataset available at \href{https://github.com/LightGal/VideoTitling}{https://github.com/LightGal/VideoTitling}}. Relatively extensive experiments on various datasets demonstrate the efficacy of the proposed method.
\end{abstract}




\maketitle

\renewcommand{\thefootnote}{\fnsymbol{footnote}}
\footnotetext[1]{These authors contributed equally to this work.}
\footnotetext[2]{Corresponding Authors.}
\footnotetext{Work was performed when S. Zhang and Z. Tan were interns at Alibaba Group.}
\renewcommand{\thefootnote}{\arabic{footnote}}
\fancyhead{}

\input{1.Introduction.tex}

\input{2.Related_Works.tex}


\input{4.Method.tex}

\input{5.Experiments.tex}

\input{6.Conclusion.tex}

\bibliographystyle{ACM-Reference-Format}
\bibliography{9.citations}

\appendix
\input{7.Appendix.tex}

%

\end{document}

%% file: 1.Introduction.tex
\section{Introduction}




Nowadays, a massive number of short videos are being generated, and short video applications play a pivotal role in the acquisition of new customers \cite{Cheng_Xiao_Li_Zhao_Wu_2019}. In e-commerce, a great many consumers upload videos in the product comment area to share their unique shopping experiences. Different from professional agency-generated videos (\eg, Ads), which may deliver a wealth of official information, consumer-generated videos depict consumers’ individual preference for the different aspects of certain products, which probably tempt others to buy the same products. Therefore, we are more interested in recommending consumer-generated videos to potential consumers. For video recommendation, it is in general appropriate to generate a descriptive and memorable title for the given video to have potential consumers know the benefit they will receive before diving into the full video contents. 



\begin{figure}[!t] \begin{center}
    \includegraphics[width=0.9\columnwidth]{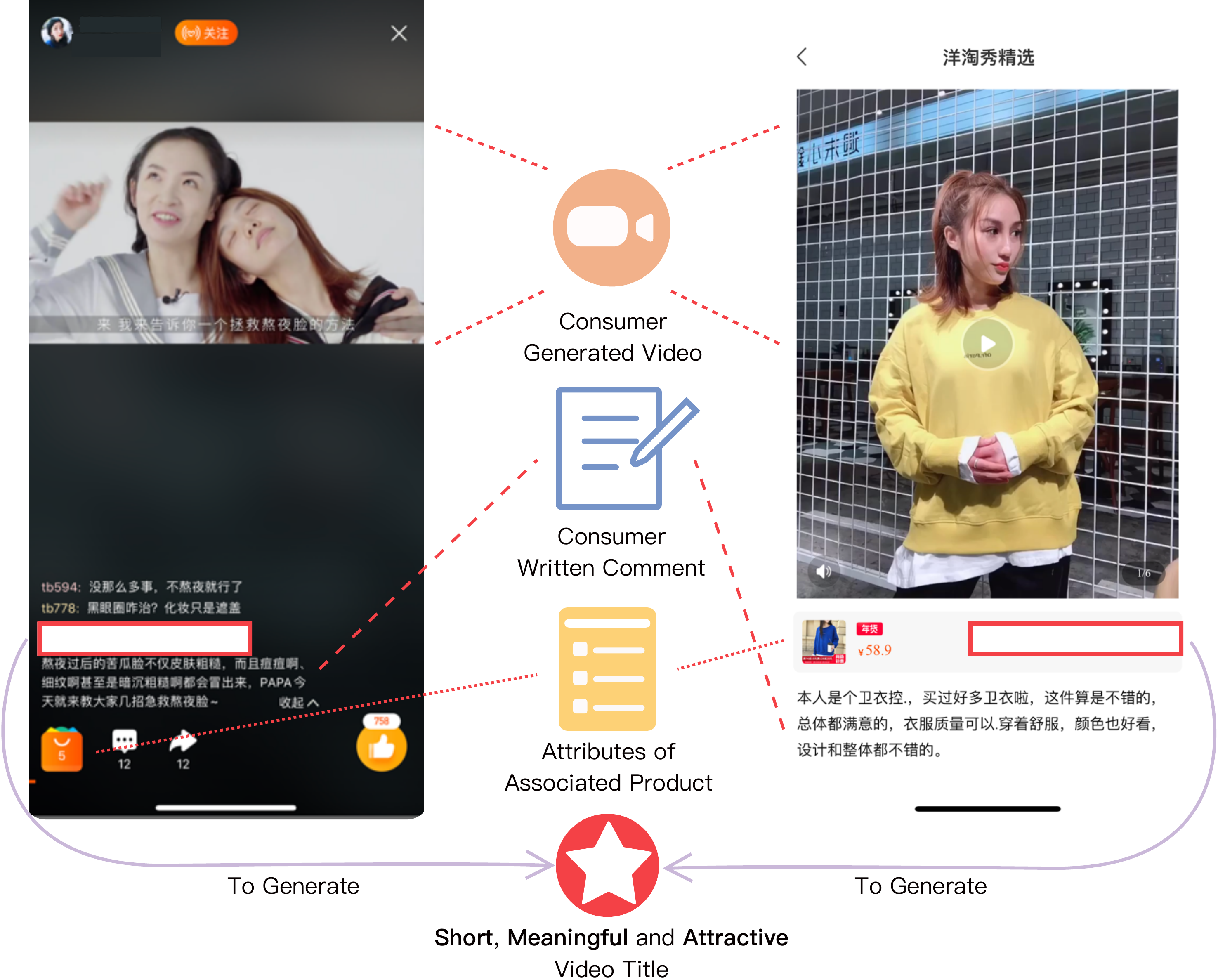}
    \caption{
    An illustration of task definition and two real application scenarios in Mobile Taobao, \ie, \textit{Trending Video Topics} (left) and \textit{Selected Buyers Show} (right). We aim to generate titles for consumer-generated videos considering the narrative comments and the attributes of associated products.
	}
\vspace{-0.6cm}
\label{fig:application}
\end{center} \end{figure}

Although the consumer-generated videos often accompany the user-written descriptive comments, these comments are infeasible to be directly taken as titles since many commentary sentences often emphasize the consumers’ intuitive perception of some shopping experience (e.g., the logistics and the services) irrelevant to the products themselves (e.g., the functionality of products). To this end, we seek an appropriate way to generate titles of consumer-generated videos from the perspective of distilling overall meaningful information depending on the following three kinds of facts: 1) the consumer-generated videos, which visually illustrate the detailed characteristics and story-line topics of products; 2) the comment sentences written by consumers, which mainly narrate the consumers’ preference to different aspects of products as well as consumers’ shopping experience; 3) the attributes of associated products, which specify the human-nameable qualities of the products (e.g.,  functionality, appearance). It can be observed that each of them contains unique and beneficial information that should not be overlooked. We name our task as product-aware \textit{video titling} ( the generation of video titles). Figure \ref{fig:application} illustrates two real application scenarios of \textit{video titling} – Trending Video Topics and Selected Buyers Show in Mobile Taobao, which is the largest online e-commerce platform in China.

To the best of our knowledge, we are the initiative to investigate such a problem in a real complex scenario. One related and general task in the literature can be video captioning. The challenging and practical nature of the task lends itself to various models \cite{Venugopalan_Rohrbach_Donahue_Mooney_Darrell_Saenko_2015,Yao_Torabi_Cho_Ballas_Pal_Larochelle_Courville_2015,Pan_Xu_Yang_Wu_Zhuang_2016,Wang_Ma_Zhang_Liu_2018}. However, these RNN-Encoder based methods have limitations. On the one hand, these methods model the video solely in the frame-level, which may be suitable for describing videos as a whole (\ie, recognizing main objects and general actions such as " a man is playing basketball") but may fail to recognize the distinguishing features of products as well as the dynamic change of fine-grained landmarks, \ie, different key-points of products. On the other hand, the original encoder-decoder framework of RNNs is incapable of systematically employing and aggregating the three kinds of information: the visual spatial-temporal dynamics in the video, the narrative description in commentary sentences and the human-nameable aspects of products.



To mitigate these problems, we propose a new learning schema named \textit{\textbf{G}raph b\textbf{a}sed \textbf{V}ide\textbf{o} \textbf{T}i\textbf{t}le g\textbf{e}nerator} abbreviated as Gavotte. Specifically, \Our is hierarchically comprised of two sub-processes, \ie, \textit{granular-level landmark modeling} and \textit{abstraction-level story-line summarization}. The granular-level sub-process represents the three kinds of information ( \ie, the consumer-generated video, the narrative commentary sentence, as well as the attributes of the associated product) as three individual graphs, in order to better capitalize on the intra-actions of granular-level cues inside three kinds of information. Furthermore, the sheer difficulty of capturing the inter-actions of different heterogeneous graphs and aggregating these graphs has not been well investigated yet in the literature. To this end, we propose a \textit{Global-local Aggregation} module enhanced with the soft attention mechanism to effectively model the inter-actions and aggregate heterogeneous graphs into a holistic graph representation. Since consumers usually customize their favorite products with the social surrounding and physical environment (e.g., lighting and decorations) in demonstration videos, it is beneficial to appropriately utilize the interactions between products and backgrounds in order to generate better titles. To capture such information in the frame-level and story topic at the video-level, the \textit{abstraction-level story-line summarization} process models the temporal structure of frames using RNNs with the guidance of granular-level features, driven by the empirical observation that the high-level semantics such as the topic and style are closely related to the local details.


To accommodate our research and applied system development, we collect a large-scale video title generation dataset, named \textit{Taobao video titling dataset} (\textbf{T-VTD}) from real-world e-commerce scenario. T-VTD contains about 90,000 <video, comment, attributes, title> quadruples, and is several orders of magnitude greater than most general video captioning datasets considering the quantity and total length of videos. For natural language data, T-VTD has an extensive vocabulary and less repetitive information. Notably, compared to captions from existing video datasets, which mostly describe the objects directly, the titles in our datasets depict different levels of information, including the fine-grained characteristics, the main category, and the overall style of the products as well as video topics. These new features of T-VTD pose many new challenges for general video captioning research and other diversified research topics. We make the desensitized version of T-VTD publicly available to promote further investigation and make our model reproducible.



In summary, the main contributions of this work are three-fold.

\begin{itemize}
	\item We devise a new learning schema for general video captioning by employing the flexible GNNs and the hierarchical video modeling processes, \ie, the \textit{granular-level landmark modeling}, and the \textit{abstraction-level story-line summarization}.
	\item We advocate investigating the real-world problem in e-commerce, named product-aware \textit{video titling}, and propose the \textit{Global-local Aggregation} module to capture the inter-actions across heterogeneous graphs and aggregate them into a holistic representation.
	\item We conduct relatively extensive experiments across various measurements, including human evaluation and online A/B test, on a large industry dataset from the Mobile Taobao. We will also release the Taobao dataset to further nourish the research community.
\end{itemize}






%% file: 2.Related_Works.tex
\section{Related Works}



\subsection{Video Captioning \& Title Generation}

For traditional video captioning, template-based methods have made substantial improvements for their ease-of-use and reliable performance \cite{Rohrbach_Qiu_Titov_Thater_Pinkal_Schiele_2013,Kojima_Tamura_Fukunaga_2002,Guadarrama_Krishnamoorthy_Malkarnenkar_Venugopalan_Mooney_Darrell_Saenko_2013}. Nevertheless, these kinds of methods highly depend on manually and carefully predefined templates. They are also limited in expression ability since only partial output (word roles, such as subject, verb, and object) is generated.

Notably, deep learning based video captioning methods obtained great success and can mitigate the problems above. Most of them rely on frame features pre-extracted by other off-the-shelf general video understanding models \cite{Feichtenhofer_Pinz_Wildes_2016,Hara_Kataoka_Satoh_2018}. Venugopalan \etal \cite{Venugopalan_Xu_Donahue_Rohrbach_Mooney_Saenko_2015} proposes to represent the video as mean-pooled frame features. More advanced methods employ the effective sequence-to-sequence encoder-decoder architecture \cite{Venugopalan_Rohrbach_Donahue_Mooney_Darrell_Saenko_2015}, hierarchical RNN design \cite{Pan_Xu_Yang_Wu_Zhuang_2016}, and soft-attention mechanisms \cite{Yao_Torabi_Cho_Ballas_Pal_Larochelle_Courville_2015}. More recently, Wang \etal \cite{Wang_Ma_Zhang_Liu_2018} proposes a reconstruction backward-flow to re-generate the input frame features both locally and globally, ensuring that the decoder hidden vectors contain the necessary video information. Livebot \cite{Ma_Cui_Dai_Wei_Sun_2019} employs a Unified Transformer Model to generate live video comments based on frames and surrounding comments. The main differences between our proposed method and the above are mainly two-fold: 1) They solely model the video in the frame-level to recognize the main object and general event for captioning. Gavotte represents landmark-level features among video frames as a video landmark graph to better capture the granular cues of products for product-aware video titling. 2) These models are incapable of modeling the comprehensive sources of facts. We propose the \textit{global-local aggregation} module to model the fine-grained inter-actions between these facts and aggregate them into a holistic representation.


To the best of our knowledge, Zeng \etal \cite{Zeng_Chen_Niebles_Sun_2016} is currently the only work that investigates video title generation, which aims to capture the most salient events in videos by proposing a highlight detector. However, this design requires highlight moments annotation for training the highlight detector, which does not apply to our task. It may not recognize the fine-grained characteristics due to the frame-level representation, and cannot model the comprehensive sources of input facts, either.

\subsection{Video Graph Representation}

To exploit special relationships besides the over-explored sequential frame dependencies, there are several lines of works representing video as graphs. Wang \etal \cite{Wang_Gupta_2018} firstly represents the objects within and across frames as a spatial-temporal graph for video action recognition. Following this, VRD-GCN \cite{Qian_Zhuang_Li_Xiao_Pu_Xiao_2019} builds a video graph using a similar setting and design the ST-GCN module to perform information propagation. AGNN \cite{Wang_Lu_Shen_Crandall_Shao_2019} views frames as nodes and models the fully-connected relationships using Graph Neural Networks. To capture the video segment level interactions, Zeng \etal \cite{Zeng_Huang_Tan_Rong_Zhao_Huang_Gan_2019} represents the pre-extracted segment proposals as the graph for action localization. To the best of our knowledge, our work is an early attempt to apply video graph representation for describing videos in natural language. We differ from the above methods by representing the granular-level landmark cues as a graph instead of the object- or frame-level information. Notably, besides the visual graph, we also represent information of other modality and other structures (\ie, the narrative description, and human-namable attributes) as graphs. We design the \textit{global-local aggregation} module to align and aggregate heterogeneous graphs into a holistic representation rather than single graph modeling.



%
%
%

%
%
%

%% file: 4.Method.tex
\begin{figure*}[t] \begin{center}
    \includegraphics[width=0.9\textwidth]{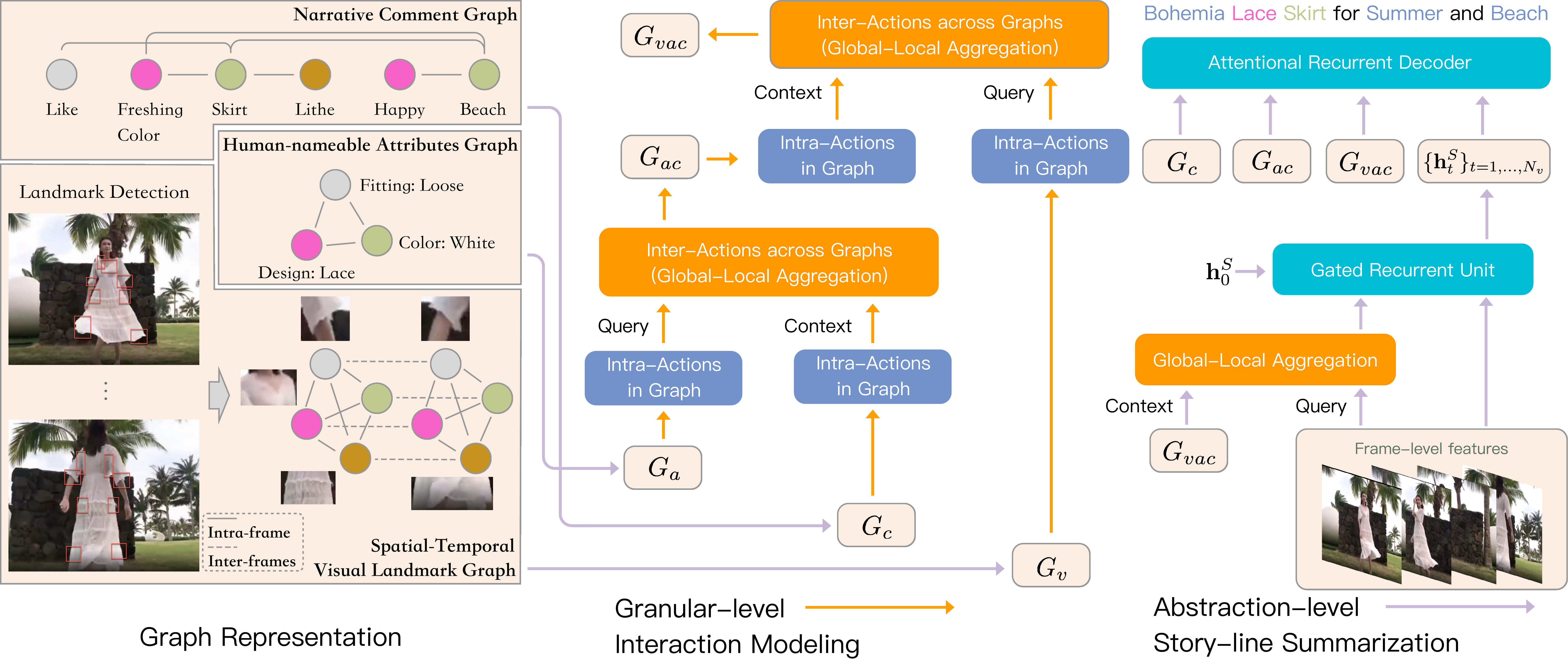}
    \caption{
    Schematic of our proposed Gavotte, comprising mainly the granular-level interaction modeling process and abstraction-level story-line summarization. The former process focuses more on recognizing the granular-level cues within different facts using graph modeling and combining heterogeneous graphs into a holistic representation, \ie, $G_{vac}$. As a necessary counterpart, the latter process uses both frame features and the aggregated graph features and is designed to figure out the product-background interaction as well as the story topic. Modules of the same kind are depicted using the same color.
	}
\vspace{-0.2cm}
\label{fig:schema}
\end{center} \end{figure*}

\section{Methods}

\begin{figure}[!t] \begin{center}
    \includegraphics[width=0.9\columnwidth]{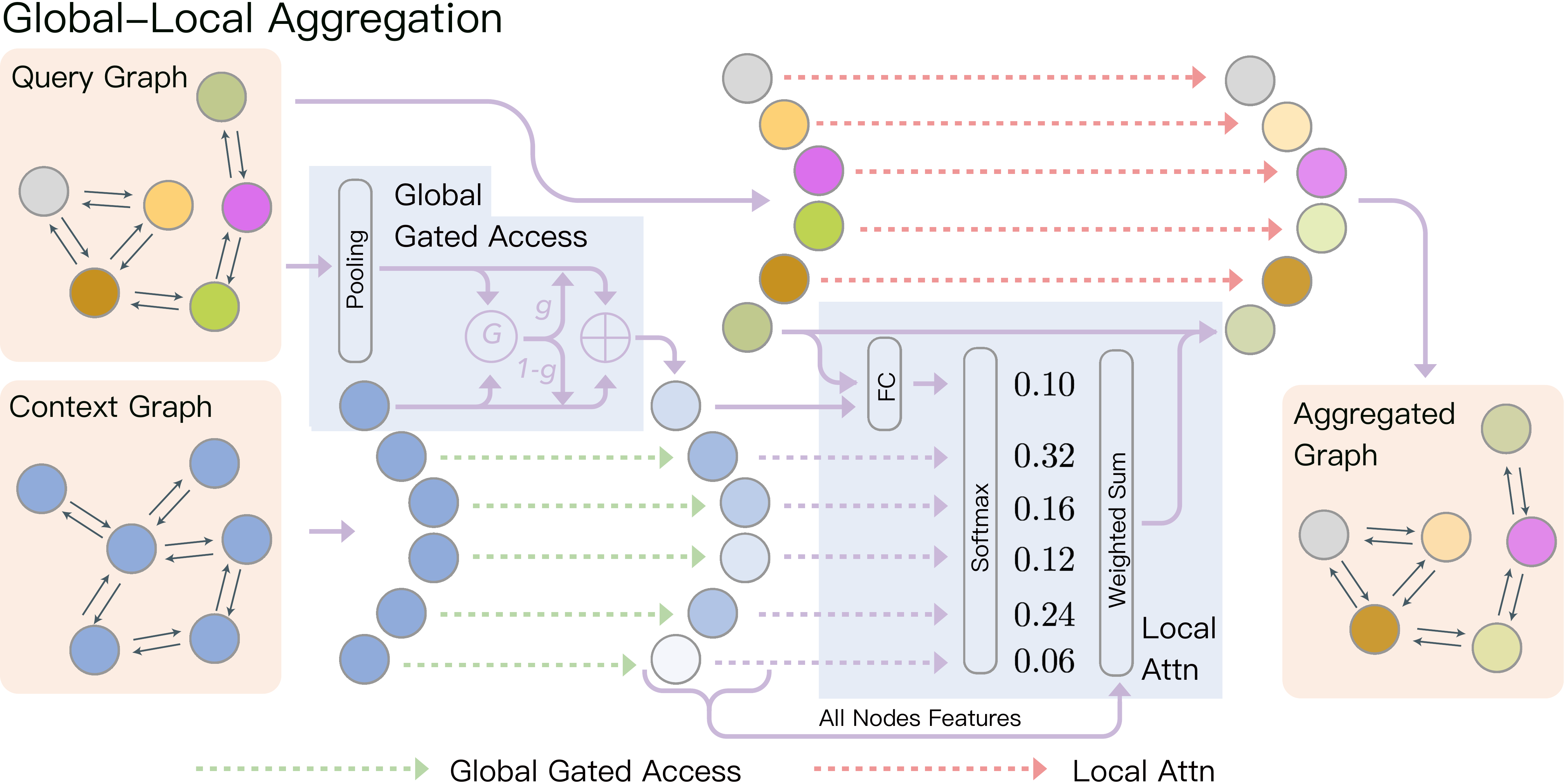}
    \caption{
    A detailed illustration of the proposed global-local aggregation module for aligning and combining heterogeneous graphs. We firstly encourage context information relevant to the global representation of the query graph (global gated access) and further attend to the context graph for each query node (local attention).
	}
\vspace{-0.4cm}
\label{fig:GLA}
\end{center} \end{figure}

\subsection{Overview}

Given a consumer-generated video comprised of $N_v$ frames, \ie, $\{x_{i}\}_{i = 1,\dots,N_v} $, a consumer-written comment of $N_w$ words, \ie, $\{w_{t}\}_{t= 1,\dots,N_w}$ and the $N_a$-sized attributes set, $\{ a_k \}_{k = 1,\dots,N_a}$, of the associated product, the goal is to generate an abstractive and descriptive video title $\{c_m\}_{m=1,\dots,N_c}$ for e-commerce mobile display. The main challenges of video titling come from the multi-modal and multi-source nature of the input information, and the sheer difficulty of capturing multi-grained cues within the video. Our proposed \model recasts the generation procedure as two heuristic sub-processes, namely \textit{granular-level interaction modeling} and \textit{abstraction-level story-line summarization} to hierarchically and comprehensively model the three kinds of information.

In the \textit{granular-level interaction modeling} process, we firstly represent the three kinds of information as individual graphs, including the video landmark graph $G_v$, the comment graph $G_c$, and the attributes graph $G_a$, separately. These graphs focus on particularly fine-grained information. We obtain the intra-actions of the granular-level cues in each graph using the flexible and effective Graph Neural Networks. To exploit the inter-actions and dependencies across graphs, we propose the \textit{global-local aggregation} module, which transforms heterogeneous graphs into an aggregated graph.

In the \textit{abstraction-level story-line summarization} process, the goal is to model the sequential structure of the video in the frame-level. This process is designed to better recognize the main topic and product-background interaction, which are essential to generate grounded and descriptive video titles. The RNN is employed here since it is capable of learning the temporal dependencies of videos



\begin{table}[t]
\centering
\setlength\doublerulesep{0.5pt}
\small
\begin{tabular}{c|l}
\hline
\hline
\multicolumn{1}{c|}{ \textbf{Notation} }        & \multicolumn{1}{l}{ \textbf{Description} }          \\
\hline \hline

$e, v, G$ & the edge, vertex and graph \\
$\mathbf{e}, \mathbf{v}$ & the edge weight / the vertex feature \\
$\mathbf{V}$ & the set of vertex features \\
$\mathbf{p}, \mathbf{r}$ & position/type embedding \\
$N$ & the number of nodes / the length of sequence/set \\
$D$ & the dimension of features \\
$\mathbf{L}_i, \mathbf{l}_{i,j}$ & landmarks feature set of the $i$th frame / the $j$th feature \\
$\mathbf{X}, \mathbf{x}_i$ & the sequence of frame features / the $i$th frame feature \\
$\mathbf{W}, \mathbf{b}$ & linear transformation / bias term \\
$\mathbf{H}, \mathbf{h}_t$ & the sequence of RNN hidden features / the feature at step $t$ \\
\hline
\end{tabular}
\caption{
    Notations
}
\vspace{-0.4cm}
\label{tab:notations}
\end{table}

\subsection{Graph Representations}


\subsubsection{Video Landmark Graph - $G_v$} 

Similar to \cite{Liu_Yan_Luo_Wang_Tang_2016}, we use landmarks to denote the salient parts of products. To effectively exploit the landmarks in each frame and across frames, we represent these landmarks of a video as a spatial-temporal graph. Previous works \cite{Wang_Gupta_2018,Wang_Lu_Shen_Crandall_Shao_2019,Zeng_Huang_Tan_Rong_Zhao_Huang_Gan_2019,Qian_Zhuang_Li_Xiao_Pu_Xiao_2019} have demonstrated the superior performance of video-graph modeling in capturing long-range temporal dependencies and the high-order relationships among the frames/objects, and we transfer this idea to the landmark-level. Specifically, we extract landmark features $\mathbf{L}_i = \{ \mathbf{l}_{i,j} \in \mathbb{R}^{D_L} \}_{j=1,\dots,N_l}$ from the $i$th frame. $N_l$ denotes how many landmarks are detected within a frame. The extraction details can be found in the appendix \ref{sub:id}. Each landmark feature $\mathbf{l}_{i,j}$ is represented as a vector of dimension $D_L$. Therefore, the video landmark graph contains $N_l \times N_v$ nodes (landmarks). We build full connections for two kinds of landmarks: 1) landmarks within the same frame, which help capture the appearance and design of the product as a whole. 2) landmarks of the same kind (\eg, collar) across frames, which help capture the dynamic change and the comprehensive characteristics of each kind of landmark. 



However, there is no notion of frame order in this schema (unlike common RNN architectures) and thus losing the temporal dependencies in the input/output. Inspired by the position embedding technique widely used in sequence learning \cite{gehring2017convolutional,lu2019vilbert}, we add a similar position-dependent signal to each landmark feature, which helps capture comprehensive landmark appearance and functionality along the timeline. We further facilitate the landmark feature by adding landmark-type dependent (such as collar and sleeve) embedding. Therefore, the node feature in $G_v$ is initialized as:
\begin{align}
	\mathbf{v}^l_{i,j} = \mathbf{l}_{i,j} + \mathbf{p}_i + \mathbf{r}_j
\end{align}
where $\mathbf{p}_i$ denotes the embedding of position $i$, \ie, frame index. $\mathbf{r}_j$ denotes the embedding of landmark-type $j$. Both $\mathbf{p}_i$ and $\mathbf{r}_j$ are learned by supervised training.

\subsubsection{Narrative Comment Graph - $G_c$} In the video titling task, the nouns and corresponding adjectives are of high importance in the narrative comment sentence. To capture these product-related words and their dependencies, we propose first to perform syntactic dependency parsing \cite{Zeman_Hajic_Popel_Potthast_Straka_Ginter_Nivre_Petrov_2018} on the comment sequence and build a narrative comment graph $G_c$ accordingly. Words having dependencies will be connected.

\subsubsection{Attributes Graph - $G_a$}  Attributes are characteristics that define a particular product (such as appearance and functionality). Each video is associated with one product in Taobao, and the seller has manually annotated many key-value pairs, such as "Color:White", for the product. In practice, we keep the essential values as the attributes set. We represent the attributes set as fully-connected graph $G_a$.


\subsubsection{Edge weights} The edges weights of all three graphs are obtained using the same schema, \ie, a learned similarity function.
\begin{align}
	\mathbf{e}_{r,s} = \mathbf{W}_d [\mathbf{v}_{r}, \mathbf{v}_{s}] + \mathbf{b}_d
\end{align}
where $\mathbf{e}_{r,s}$ is a scalar, representing the edge weight of node $v_r$ and $v_s$. $\mathbf{W}_d$ is a matrix denoting the similarity measurement and $\mathbf{b}_d$ is a bias term. A schematic graph representation example for these three kinds of graph is shown in Figure \ref{fig:schema}.

\subsection{Granular-level Interaction Modeling}



\subsubsection{Information propagation}

As a common practice, we leverage GNNs (Graph Neural Networks) as a trainable schema to aggregate information in local neighborhood nodes and exploit the rich information inherent in graph structure for each graph. Specifically, given the previous feature $\mathbf{v}_i$ of node $v_i$, the updated feature $ \mathbf{\bar v}_i \in \mathbb{R}^{D_{out}}$ can be computed as:
\begin{align}
	& \mathbf{\bar v}_i^g = \sigma(\mathbf{W}_r^g \mathbf{v}_i + \mathbf{W}_n^g \ \mathrm{MAX}(\{\mathbf{e}_{i,j} * \mathbf{v}_j, j \in \mathcal{N}(i)\})  + \mathbf{b}_n^g) \label{eq:gate_gnn} \\
	& \mathbf{\bar v}_i = \mathbf{\bar v}_i^g * (\mathbf{W}_r^h \mathbf{v}_i + \mathbf{W}_n^h \ \mathrm{MAX}(\{\mathbf{e}_{i,j} * \mathbf{v}_j, j \in \mathcal{N}(i)\})  + \mathbf{b}_n^h) \label{eq:hidden_gnn}
\end{align}
where $\mathbf{W}_r^g, \mathbf{W}_n^g, \mathbf{W}_r^h, \mathbf{W}_n^h \in \mathbb{R}^{D_{out} \times D_{in}}$ are learnable linear transformations used to project the root feature and neighbors features into a joint space. $\mathbf{b}_n^g, \mathbf{b}_n^h \in \mathbb{R}^{D_{out}}$ are bias terms. $D_{in}$ and $D_{out}$ are the dimensions of the original node feature and the updated node feature. $\mathcal{N}(i)$ denotes the neighbors index set of node $v_i$. We adopt the element-wise max function $\mathrm{MAX}$ for the empirical effectiveness over the others, such as the element-wise average. $\sigma$ is the element-wise sigmoid function, and $*$ denotes the element-wise multiplication. Equation \ref{eq:gate_gnn} aims to obtain a gate vector $\mathbf{\tilde v}_i^g$, which is designed to control how much information is needed for each position in $\mathbf{\bar v}_i$ when updating. In our experiment, this design performs consistently better than the vanilla design without gate and GCNs \cite{kipf2016semi}, which is widely used in recent video graph modeling. 





\subsubsection{Global-local Aggregation} \label{sec:GAL}

To capture the inter-actions across heterogeneous graphs in an end-to-end manner, we propose the \textit{global-local aggregation} module, termed $GLA$. Given the query graph $G_q = \{v^q_i\}_{i=1,\dots,N_q}$, and the context graph $G_e = \{v^e_i\}_{i=1,\dots,N_e}$, we aim to obtain an aggregated graph $G_{agg}$ with the same structure as $G_q$ using two sub-modules, \ie, \textit{global gated access}, and \textit{local attention}. We use $\mathbf{v}^q \in \mathbb{R}^{D_q}$ and $\mathbf{v}^e \in \mathbb{R}^{D_e}$ to denote the dense representation for the query node $v^q$ and the context node $v^e$.

\vpara{Global gated access} We first obtain the global representation $\mathbf{v}^{q*}$ of the query graph by global-average pooling, and then encourage relevant information in each context node $\mathbf{v}^e$ and suppress irrelevant ones using a gate function. The query-aware context feature $\mathbf{v}_j^{eq} \in \mathbb{R}^{D_{eq}}$ can be obtained by:
\begin{align}
	\mathbf{v}^{q*} &=  1/N_q \sum_{i=1}^{N_q} \mathbf{v}^q_i \\
	\mathbf{g}_j &= \sigma( \mathbf{W}_g[  \mathbf{v}^{q*},\mathbf{v}^e_j] + \mathbf{b}_g) \\
	\mathbf{v}_j^{eq} &= (1 - \mathbf{g}_j) * (\mathbf{W}_q\mathbf{v}^{q*} + \mathbf{b}_q) + \mathbf{g}_j * (\mathbf{W}_e \mathbf{v}_j^e  + \mathbf{b}_e)
\end{align}
where $\mathbf{W}_g \in \mathbb{R}^{1 \times (D_q+D_e)}$, $\mathbf{W}_q \in \mathbb{R}^{D_{eq} \times D_q}$ and $\mathbf{W}_e \in \mathbb{R}^{D_{eq} \times D_e}$ are linear transformation matrices. $\mathbf{W}_g$ models the relevance between the global query graph representation $\mathbf{v}^{q*}$ and one context node $\mathbf{v}^e_j$. $\sigma$ denotes the sigmoid function. $\mathbf{g}_j \in [0,1]$ shows the relevance. $\mathbf{W}_q$ and $\mathbf{W}_e$ project the original representations into a query-context joint space (global). Intuitively, a larger $\mathbf{g}_j$ will encourage globally relevant information within the context graph. Smaller $\mathbf{g}_j$ will suppress the irrelevant and consider more global query information.


\vpara{Local attention} While the above process focuses on recognizing globally relevant information in the context graph (globally), the local attention sub-module aims to further filter important features related to each individual query node (locally). We perform node-level additive attention \cite{Bahdanau_Cho_Bengio_2015} which allows better selectivity in distilling relevant and necessary information. Specifically, we compute the final aggregated node vector $\mathbf{v}^{agg}_i$ as the following:
\begin{align}
	\mathbf{o}_{i,j} &= \tanh(\mathbf{W}_o [\mathbf{v}_i^q, \mathbf{v}^{eq}_j] + \mathbf{b}_o) \label{eq:a1} \\
	\mathbf{\bar o}_{i,j} &= \frac{  \exp(\mathbf{W}_a \mathbf{o}_{i,j}) }{\sum_k \exp( \mathbf{W}_a \mathbf{o}_{i,k}) } \label{eq:a2} \\
	\mathbf{v}^{agg}_i &= \mathbf{v}_i^q +  \sum_{j=1}^{N_e} \mathbf{\bar o}_{i,j} * \mathbf{v}^{eq}_j \label{eq:a3}
\end{align}
where $\mathbf{W}_o \in \mathbb{R}^{D_o \times (D_q+D_{eq})}$ and $\mathbf{W}_a \in \mathbb{R}^{1 \times D_o}$ jointly model the node-level relevance between one query node $\mathbf{v}_i^q$ and one updated context node $\mathbf{v}^{eq}_j$. $\mathbf{\bar o}_{i,j} \in [0,1]$ indicates the relevance score. Finally, all locally relevant information have been distilled and summed up to obtain $\mathbf{v}^{agg}$.


\subsubsection{Aggregating three graphs} We progressively perform information propagation for three individual graphs as well as the aggregated graphs, which are obtained by the global-local aggregation module, as depicted in Fig \ref{fig:GLA}. In practice, we firstly aggregate the narrative comment graph and the attribute graph into the attribute-comment (AC) graph $G_{ac}$. We use the attributes graph as the query graph and distill relevant (globally and locally) information from the narrative comment graph due to the observation that attributes are always more systematic and meaningful. We finally obtain the video-attribute-comment (VAC) graph $G_{vac}$ by viewing the video landmark graph as query and $G_{ac}$ as the context graph since video can be the critical part for \textit{video titling}. The VAC graph conveys the necessary granular-level cues, especially the landmark characteristics of the product, \eg, "\texttt{V-collar}" or "\texttt{bat sleeve}".

\subsection{Abstraction-level Story-line Summarization}


The above \textit{granular-level interaction modeling process} appropriately considers granular-level details of each product. As aforementioned, consumers usually customize their favorite products with the social surrounding and physical environment (\eg, lighting, and decorations) in the demonstration, in order to learn the background-product interaction and generate the story-line video topic, we design the \textit{abstraction-level story-line summarization} process.

Formally, given the aggregated graph $G_{vac}$ and the frame-level features $\mathbf{X}=\{\mathbf{x}_t\}_{t=1,\dots,N_v}$  (the detailed extraction process can be found in appendix \ref{sub:id}), we firstly use $\mathbf{X}$ as the query matrix, the nodes features of $G_{vac}$, \ie, $\mathbf{V}_{vac}$, as context matrix and perform global-local aggregation (GLA) to obtain the aggregated feature sequence:
\begin{align}
	\mathbf{\bar X} = GLA(\mathbf{X}, \mathbf{V}_{vac})
\end{align}
where $GLA$ performs the same operations as illustrated in Section \ref{sec:GAL}. This operation is intuitive under the empirical observation that the high-level picture should be consistent with the local design. Finally, we encapsulate the Gated Recurrent Unit (GRU) \cite{Cho_Van_Merrinboer_Bahdanau_Bengio_2014} to model the narrative structure of the frame feature sequence: 
\begin{align}
	\mathbf{\tilde{x}}_t &= [\mathbf{\bar x}_t, \mathbf{x}_t] \\
	\mathbf{h}_{t}^S &= \mathrm{GRU}(\mathbf{h}_{t-1}^S, \mathbf{\tilde{x}}_t)
\end{align}
We concatenate the globally-locally aggregated feature $\mathbf{\bar x}_t \in \mathbf{\bar X}$ and frame feature $\mathbf{x}_t$ as the input of GRU at time step $t$ (The details of GRU can be found in the original paper\cite{Cho_Van_Merrinboer_Bahdanau_Bengio_2014}). We keep the hidden features $\mathbf{H}^S = \{\mathbf{h}_t^S\}_{t=1,\dots,N_v}$ of all steps for decoding.
\subsection{Decoder}

As a common practice, we employ the RNN as the decoder. Differently, we initialize the state $\mathbf{h}^D_0$ as the final state of the story-line summarization RNN, \ie, $\mathbf{h}_0^D = \mathbf{h}_{N_v}^S$. In the decoding stage, we consider information within $G_{vac}$, $G_{ac}$ and $G_{c}$ graphs as well as the hidden features of the story-line reasoning RNN, $\mathbf{H}^S$. $G_{vac}$ is obtained by using the visual landmark feature as a query and thus containing more visual information. Therefore, we consider graphs $G_{ac}$ and $G_{c}$ as necessary complements to incorporate the granular-level cues within the product attributes and the narrative comment. Formally, at decoder step $m$, we use the decoder hidden state $\mathbf{h}_m^D$ as query and apply the aforementioned local attention mechanism $f_{la}$ to distill relevant information and generate titles depending on the four kinds of information, \ie, $G_{vac}$, $G_{ac}$, $G_{c}$ and $\mathbf{H}^S$:
\begin{align}
	\mathbf{x}^D_{m} = [f_{la} (\mathbf{h}_{m}^{D}, \mathbf{V}_{vac}), 
		f_{la} (\mathbf{h}_{m}^{D}, \mathbf{V}_{ac}), f_{la} (\mathbf{h}_{m}^{D}, \mathbf{V}_{c}), 
		f_{la} (\mathbf{h}_{m}^{D}, \mathbf{H}^{S})]
\end{align}
Where $\mathbf{V}$ denotes all nodes features of a specific graph. We further concatenate the context feature $\mathbf{x}^D_{m}$ and the embedding $\mathbf{\hat c}_m$ of previously predicted word ${\hat c}_m$ as the input of decoding RNN. Specifically, we use GRU for its efficiency:
\begin{align}
	\mathbf{h}_{m+1}^{D} = \mathrm{GRU}(\mathbf{h}_{m}^{D}, [\mathbf{\hat c}_m, \mathbf{x}^D_{m}])
\end{align}


\subsection{Learning Objectives}


\vpara{Cross-Entropy Loss} Our model minimizes the cross-entropy loss at each decoder step and we apply teacher forcing \cite{Williams_Zipser_1989} during training. With the reference video title denoted as $C^R = \{ c_m^R \}_{m=1,\dots,N_c}$, the cross-entropy loss can be defined as:
\begin{align}
	\mathcal{L}_{ce}=-\frac{1}{N_c} \sum_{m=1}^{N_c} \log f_P\left(c_m^R\right)
\end{align}
where $f_P\left(c_m^R\right)$ is the predicted probability of word $c_m^R$ at timestamp $m$. The predicted probability distribution $P_m^O \in \mathbb{R}^{N_O}$ on the whole vocabulary $O$ is obtained by applying one fully-connected layer $\mathbf{W}_p$ (with bias term $\mathbf{b}_p$) on the decoder hidden vector $\mathbf{h}_m^D$, followed by a softmax layer.
\begin{align}
	P_m^O = \textup{softmax} (\mathbf{W}_p \mathbf{h}_m^D + \mathbf{b}_p)
\end{align}

\vpara{Generation Coverage Loss} Despite the commonly used maximum-likelihood cross-entropy loss, we incorporate a probability-based loss similar to the coverage loss used in Text Summarization \cite{See_Liu_Manning_2017}. Instead of penalizing repetitively attending to the same location based on accumulative attention weights in the coverage loss, we directly suppress the repetitive generated words based on accumulative probability distribution. This inductive bias is reasonable due to the findings that repetitive words are mostly not expressive and attractive for \textit{video titling}. Technically, we keep a generation coverage vector $\mathbf{\varsigma}$ to store the accumulative probability distribution:
\begin{align}
	\mathbf{{\varsigma}}_m = \sum_{m^\prime=1}^{m-1} P_{m^\prime}^O
\end{align}
To directly penalize the repetitive predicted words, our model minimizes the following loss:
\begin{align}
	 { \mathcal{L}_{gc} }= \sum_{m} \sum_{i} \min \left(\mathbf{{\varsigma}}_{m,i}, P_{m,i}^O\right)
\end{align}
Intuitively, when some word is predicted with high probability in the past, the corresponding value, \ie, \ $\mathbf{{\varsigma}}_{m,i}$, will be large. If the word probability in the the current prediction, $P_{m,i}^O$, is also large, the loss will become large. The final loss can be written as:
\begin{align}
	\mathcal{L} = \mathcal{L}_{ce} + \lambda_{gc} \mathcal{L}_{gc}
\end{align}

%% file: 5.Experiments.tex
\section{Experiments}

\subsection{Data Collection} \label{Data Collection}

We collect a large-scale industrial \textit{video titling} dataset, named \textit{Taobao video titling dataset} (\textbf{T-VTD}) from real-world e-commerce scenario. T-VTD are comprised of 90,000 <video, comment, attributes, title> quadruples and each contains a product-oriented video stream from Mobile Taobao, a narrative comment uploaded alongside the video, human-nameable attributes from the associated product and a concise human-written video title. The basic statistics and comparison with benchmark video captioning (or related) datasets can be found in Appendix \ref{sub:data}.

\begin{table*}[t]
\centering
\setlength{\tabcolsep}{7.5pt}
\setlength\doublerulesep{0.5pt}
\begin{tabular}{l|cccc|cccc}
& \multicolumn{4}{c}{T-VTD} & \multicolumn{4}{c}{Other Categories}  \\
\multicolumn{1}{c|}{Method}     & BLEU-4         & METEOR   &ROUGE\_L          & CIDEr       & BLEU-4         & METEOR   &ROUGE\_L          & CIDEr          \\
%
\hline \hline
M-MPLSTM          & 0.95 & 10.43 & 15.47 & 32.65   & 1.17          & 7.58          & 13.12           & 25.75          \\
M-S2VT   & 1.52 & 11.64 & 18.03 & 41.74  & 1.70   & 9.49   & 17.26   & 39.35 \\
M-HRNE   & 1.70 & 12.11 & 18.85 & 44.53   & 1.66   & 9.40   & 17.10   & 39.36 \\
M-SALSTM    & 2.01 & 12.95 & 19.81 & 48.63   & \textbf{2.33}   & 10.65   & 18.42   & 47.01
        \\
M-RecNet    & 2.00 & 12.87 & 20.33 & 50.01  & 2.24   & 10.62   & 19.18   & 47.81
        \\
M-LiveBot   & 1.86 & 12.65 & 19.56 & 47.02 & 1.86   & 9.89   & 16.93   & 40.81
        \\
\hline \hline
\Our  & \textbf{2.28}  & \textbf{13.58}  & \textbf{21.33} & \textbf{54.14} & 2.27 & \textbf{10.99} & \textbf{19.38}  & \textbf{49.13} \\
\hline
\end{tabular}
\caption{
     Performance of gavotte and various re-implemented state-of-the-art methods (for a fair comparison) in terms of four frequently used reference-based metrics. We conduct experiments on the released T-VTD (left) and one internal dataset without comprising clothes samples (right).
}
\label{tab:eval_caption}
\vspace{-0.6cm}
\end{table*}
%
%
\subsection{Evaluation Metrics}

To obtain a fair and comprehensive comparison between our methods and other state-of-the-arts, we evaluate our method in terms of various metrics, including reference-based metrics and retrieval-based metrics.

\vpara{Reference-based metrics} Following previous works in the field of video captioning, we capitalize on four reference-based metrics, BLEU4 \cite{papineni2002bleu}, METEOR \cite{banerjee2005meteor}, ROUGE\_L \cite{lin2004rouge}, CIDEr \cite{vedantam2015cider}, to quantify the 4-gram precision, the stemming and synonymy matching, the longest common sequence overlap and the human-like consensus between the generated video titles and the references, respectively.

\vpara{Retrieval-based metrics} Despite the well-known advantages of the reference-based metrics, they are arguably not sufficient for our task because of the potentially large number of plausible solutions and the only-one reference for measurement. To this end, we adopt the retrieval-based evaluation protocol proposed by Abhishek \etal \cite{Das_Kottur_Gupta_Singh_Yadav_Moura_Parikh_Batra_2017}, which returns a sort of candidate titles based on the log-likelihood score rather than directly compare the generated title with the reference. In a similar spirit, for each test sample, we select candidate titles, including the title of the test sample, 50 plausible titles, 20 popular titles, and 29 randomly sampled titles. The plausibility is measured by the angular similarity of tf-idf dense representations of attributes sets. For popularity, since there are few repetitive titles in our dataset unlike in \cite{Das_Kottur_Gupta_Singh_Yadav_Moura_Parikh_Batra_2017}, we choose the top 20 titles with the most number of similar neighbors. We also use angular similarity, which distinguishes almost identical vectors much better, of tf-idf dense representations.



\subsection{Competitors}

To the best of our knowledge, there are no methods doing precisely the same task as ours before. We re-implement and make necessary modifications (for a fair comparison) to various publicly available video captioning methods (or related) as our competitors. We mainly add separate encoders for additional inputs, \ie, the human-nameable attributes, and narrative comment, and concatenate the encoder outputs. In detail, we adopt:
\begin{itemize}
	\item \textit{M-MPLSTM}. M-MPLSTM adds two average-pooling encoders over MPLSTM \cite{Venugopalan_Xu_Donahue_Rohrbach_Mooney_Saenko_2015}. 
	\item \textit{M-S2VT}. M-S2VT replaces the naive mean-pooling strategy in M-MPLSTM by LSTM encoders.
	\item \textit{M-HRNE} We re-implement the HRNE model \cite{Pan_Xu_Yang_Wu_Zhuang_2016} by modeling each input modality with a separate hierarchical encoder.
	\item \textit{M-SALSTM} Based on M-S2VT, M-SALSTM further leverages the effective soft attention mechanism by attending to all three encoded feature sets in each decoding step.
	\item \textit{M-RecNet} We re-implemented RecNet \cite{Wang_Ma_Zhang_Liu_2018} over M-SALSTM by reconstructing the initial features of all inputs. 
	\item \textit{M-LiveBot} We modify the LiveBot \cite{Ma_Cui_Dai_Wei_Sun_2019} by adding an additional transformer encoder to model the attributes.
\end{itemize}

%
%
%
%
%
%
%

\subsection{Performance Analysis}

\vpara{Quantitative Results} In summary, the results on both the reference-based metrics and the retrieval-based metrics consistently indicate that our proposed method achieves better results against various Video Caption methods, including basic RNN models (M-MPLSTM and M-S2VT), attention-based RNN approach (M-SALSTM and M-RecNet) and transformer-based architecture (M-LiveBot).


\begin{table}[!t]
\centering
\setlength{\tabcolsep}{4.5pt}
\setlength\doublerulesep{0.5pt}
\begin{tabular}{l|ccccc}


\multicolumn{1}{c|}{Models} &    R@1 $\uparrow$      &   R@5 $\uparrow$   &    R@10  $\uparrow$  & MR $\downarrow$ &  MRR  $\uparrow$   \\
\hline \hline

M-MPLSTM            & 11.1          &    34.6        &    46.3 & 23.48  & 0.230               \\
M-S2VT  & 21.3   & 52.1    &   67.6 & 12.23 & 0.362  \\
M-HRNE     &  22.6  & 54.0    & 68.3  & 11.98 & 0.375
        \\				
M-SALSTM  &  25.9   & 54.5    & 67.84  & 12.16 & 0.397
\\
M-RecNet   &  26.6  & 56.4    & 69.1  & 12.04 & 0.408
\\
M-LiveBot   &  19.4  & 44.8    & 58.0  & 17.08 & 0.320

\\
\hline
\hline				
\Our     &  \textbf{26.8}  & \textbf{58.2}    & \textbf{71.0}  & \textbf{11.614} & \textbf{0.415}
\\

\hline

\end{tabular}
\caption{
Performance analysis using the retrieval-based metrics on T-VTD.
}
\label{table:Ranking}
\vspace{-0.6cm}
\end{table}

\begin{figure*}[t] \begin{center}
    \includegraphics[width=0.9\textwidth]{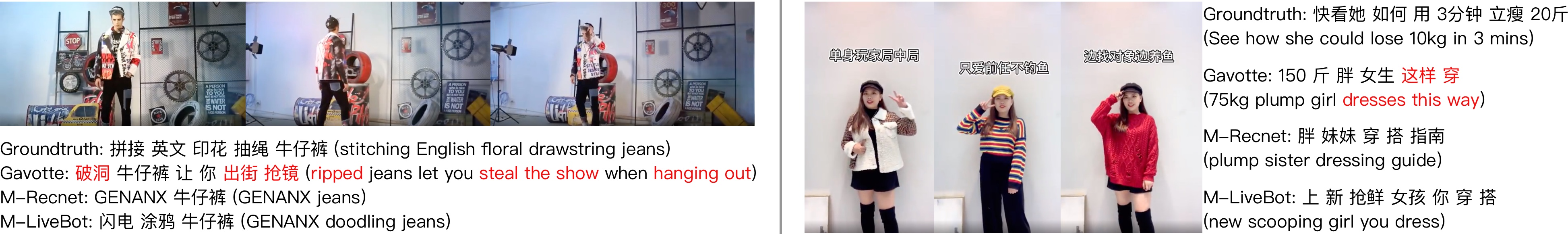}
    \caption{
    	Title examples generated by Gavotte, M-RecNet, and M-Livebot with sampled frames of corresponding input videos. Results show that our model can generate more fluent and meaningful titles with attractive buzzwords like "\texttt{style-freshing}".
	}
\vspace{-0.0cm}
\label{fig:case}
\end{center} \end{figure*}


Analogously to gains observed in other natural language generation tasks, the RNN based encoder (M-S2VT) is much better than the simple mean-pooling strategy (M-MPLSTM) in our task. A simple hierarchical design (M-HRNE) can lead to minor performance boost. Notably, attention-based architectures (M-SALSTM, M-RecNet, M-LiveBot, Gavotte) achieve the first-tier results due to its superior power on selecting important features and suppressing unnecessary ones. This ability is essential for our task since the three kinds of facts (\ie, the consumer-generated video, narrative comment, and seller-generated attributes) can be noisy and unreliable in the real scenario. The reconstruction flow in M-RecNet guarantees that the decoded representation contains summarized information of the input and thus improving the M-SALSTM. The transformer-based M-LiveBot can not achieve competitive results than the M-SALSTM and M-RecNet. We relate the results to the inferior capacity of transformer-decoder than attentional RNN decoder \cite{Chen_Firat_Bapna_Johnson_Macherey_Foster_Jones_Schuster_Shazeer_Parmar_2018} and its heavy dependence on pre-trained word embedding to avoid over-fitting. Our method outperforms the M-RecNet (the best among competitors) by almost 8.26\% considering the CIDEr metric. We attribute the substantial improvements to the following reasons: 1) The graph representation can model special relationships such as the interactions of landmarks compared to the vanilla sequential dependency. 2) The proposed \textit{global-local aggregation} module can better capture the inter-actions across heterogeneous graphs, resulting in an aligned and aggregated modeling. 3) The hierarchical design of granular-level interaction modeling and abstraction-level story-line summarization is essential for obtaining a comprehensive understanding of the video. 



To verify the model capacity on non-clothes categories dataset, we conduct an additional experiment, of which the results are shown in Table \ref{tab:eval_caption}. This internal dataset contains 90,000 videos belonging to three categories, \ie, toys, makeups, and kitchen tools. Products of these categories do not have well pre-defined landmarks like clothes (\eg, collar, and sleeve). We choose to first locate the presence of product with a bounding box and view the equally divided 3x3 areas of the bounding box as landmarks. We represent these nine landmarks using average-pooled features. This setting can be viewed as an approximation to the setting for clothes. The improvements over other competitors are comparatively smaller than the improvement on T-VTD. This is a reasonable result due to the coarser landmark feature extraction. Despite these disadvantages, our method achieves the best result on most measurements, which further demonstrate the merit of our design.

\begin{table}[!t]
\centering
\setlength{\tabcolsep}{5.5pt}
\setlength\doublerulesep{0.5pt}
\begin{tabular}{l|ccc}


\multicolumn{1}{c|}{Models} &    Fluency      &   Diversity    &    Overall Quality  \\
\hline \hline

M-RecNet   &  3.671  & 3.497    & 3.310
\\
M-LiveBot   &  3.901  & 3.613    & 3.407
\\
\hline
\hline
\Our   & \textbf{3.997}  & \textbf{3.671}    & \textbf{3.536} 
\\

\hline

\end{tabular}
\caption{
Human evaluation results across three aspects.
}
\vspace{-0.6cm}
\label{table:HumanEval}
\end{table}


\vpara{Human Evaluation} To obtain a more reliable and precise measurement of the generation results, we conduct human judgments concerning the following three aspects \cite{Li_Monroe_Shi_Jean_Ritter_Jurafsky_2017,Chen_Lin_Zhang_Yang_Zhou_Tang_2019}: Fluency, Diversity and Overall Quality. \textit{Fluency} is designed to grade how fluent the generated titles are. The score of \textit{Diversity} reflects whether the model generates refreshing content and is expected to be low when the titles are repetitive and dull. \textit{Overall Quality} is intended to measure whether the titles are consistent with the three kinds of information, \ie, video, comment, and attributes. The score range for each aspect is from 1 to 5. We consider M-RecNet and M-LiveBot as the representatives of RNN-based methods and transformer-based methods, and thus as competitors. We ask crowd workers in Taobao to grade the generation results of randomly sampled 1000 instances from the test set. Each instance contains the video link, the user-written comment, the attributes of the associated product, and generation results from three models. Workers are required to first watch the video before reading the other inputs and titles. According to the evaluation results shown in Table \ref{table:HumanEval}, Gavotte can generate more fluent and more diversified titles than the other two competitors. As for the measurement of faithfulness to inputs and some grounded facts (overall quality), Gavotte achieves a clear improvement over M-RecNet and M-LiveBot by +0.129 and +0.226 respectively. An interesting finding is that M-LiveBot performs much better than M-RecNet in human evaluation. By qualitatively comparing generation samples, we find that the titles generated by M-RecNet are short and accurate, which can be beneficial for automatic measurements. In contrast, a self-attention based model (M-Livebot) generates longer titles with more comparatively uncommon words, which can be more attractive for human judgers.

%

\vpara{Qualitative Results} Figure \ref{fig:case} shows some title generation samples from Gavotte and the other two competitors, \ie, M-RecNet, and M-Livebot. Similar to the results in human evaluation, Gavotte can generate more fluent and attractive titles. Specifically, while the title of M-Recnet in the first case is less informative and the title of M-LiveBot in the second case is unfinished or broken, Gavotte generates smooth and meaningful title with the popular buzzwords -- "\texttt{steal the show}". The results further demonstrate that Gavotte can recognize granular-level details like "\texttt{ripped}", clothes-level design like "\texttt{jeans}", frame-level product-background interaction effect like "\texttt{steal the show}", and video-level story-line topic like "\texttt{dresses this way}".

%
%

\subsection{Ablation Studies}

To better understand the behavior of different modules in our model, we surgically and progressively remove several components, including the \textit{abstraction-level story-line summarization} process (HSR), the \textit{gated global access} (GGC) and the \textit{local attention} (LA) mechanism in global-local aggregation. The results are shown in Table \ref{table:ablation}. Overall, removing any of the components leads to a performance drop, which verifies the effectiveness of these components. Removing HSR will result in a failure to recognize the background, the overall style, and the story-line topic, and the metric scores decrease accordingly. During the graphs aggregation process, failing to access the global picture will result in a performance drop (from 53.06 to 52.57 considering the CIDEr metric). Notably, further removing the LA module means the total loss of the graph inter-action process. The obvious performance gap between the LA-removed model and the previous model demonstrates that the global-local aggregation module, especially the local attention sub-module, is an essential part for this task. In addition, this structure is similar to the M-SALSTM architecture except for the graph representation and GNN based encoder. The improvement over the M-SALSTM indicates the merit of leveraging both the flexible GNNs and granular-level cues within different kinds of information.


\begin{table}[!t]
\centering
\setlength{\tabcolsep}{2.5pt}
\setlength\doublerulesep{0.5pt}
\begin{tabular}{l|cccc}


  \multicolumn{1}{c|}{Models}    & BLEU-4         & METEOR   &ROUGE\_L          & CIDEr       \\
\hline \hline

 \Our         & 2.28     &    13.58     &    21.33 & 54.14             \\

\hline 
 \ \  - HSR  &  2.20  & 13.32    & 21.02  & 53.06
\\
  \ \ \ \  - GGA   &  2.13  & 13.24    & 20.89  & 52.57 
\\
  \ \ \ \ \ \  - LA   &  2.06  & 13.33    & 20.61  & 51.48 
\\
\hline

\end{tabular}
\caption{
Model ablations by progressively removing three components.
}
\vspace{-0.4cm}
\label{table:ablation}
\end{table}



\subsection{Online A/B Test}


We deploy our model Gavotte on the \textit{Guess You Like} scenario in mobile Taobao. Originally, the titles for recommended consumer-generated videos are generated as the truncated consumer-written comment. We conduct online tests on this baseline method and Gavotte under the framework of the A/B test. The testing set contains 10,000 video samples, which has no overlap with T-VTD. The count of total page views is around 400,000, and the number of all unique visitors is about 100,000. We keep the network traffic for both methods near the same. Gavotte improves the click-through-rate by +9.90\% compared to the baseline. These results demonstrate that our model can generate meaningful and attractive titles, which further improve the performance of video recommendation.

%% file: 6.Conclusion.tex
\section{Conclusion}

In this paper, we propose a comprehensive information integration framework, named Gavotte, to generate descriptive and attractive titles for consumer-generated videos in e-commerce. We firstly represent three kinds of information (\ie, the consumer-generated video, the narrative comment sentences, and the human-nameable attributes of the associated products) as graphs. Then we perform \textit{granular-level landmark modeling} to exploit both the intra-actions within each graph using flexible GNNs and the inter-actions across graphs using proposed \textit{global-local aggregation} module. This schema is designed to figure out the granular-level characteristics, such as the appearance of product landmarks, within different kinds of information, and aggregate them into a holistic representation. We further employ \textit{abstraction-level story-line summarization} to capture the product-background interactions in the frame-level as well as the story-line topic in the video-level. As far as we know, this is the first piece of work that explores video graph modeling for general video captioning. We collect and release a corresponding large-scale dataset for reproduction and further research. The consistent quantitative improvement across various metrics reveal the effectiveness of the proposed method.

\section{ACKNOWLEDGMENTS}

The work is supported by the NSFC (61625107), Zhejiang Natural Science Foundation (LR19F020006), Fundamental Research Funds for the Central Universities (2020QNA5024, 2018AAA0101900), and a research fund supported by Alibaba.

%% file: 7.Appendix.tex
\section{Appendix}
\label{sec:appendix}

\subsection{Details on the Experimental Setup and Hyperparameters}





\begin{figure}[!h] \begin{center}
    \includegraphics[width=\columnwidth]{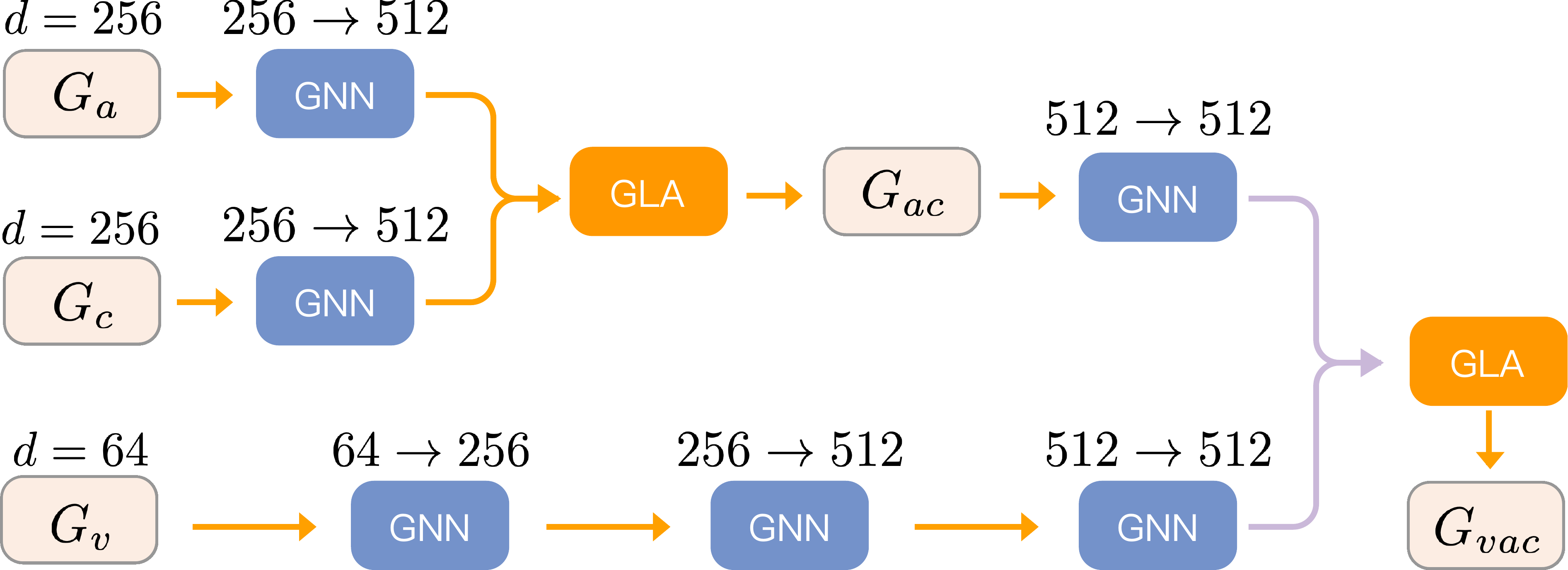}
    \caption{
    Architecture and parameter configuration of granular-level interaction modeling process.
	}
\label{fig:granular-level}
\end{center} \end{figure}

\vpara{Hyper-parameter Configuration} The parameter setup of \textit{granular level interaction modeling} process is shown in Figure \ref{fig:granular-level}. During training, the batch size is set to 64 and we use Adam optimizer \cite{Kingma_Ba_2015} with the setting $\beta _ { 1 } = 0.9 , \beta _ { 2 } = 0.999,  \mathrm{weight decay} = 1 \times 10 ^ {-4} \text { and } \epsilon = 1 \times 10 ^ { - 8 }$. Learning rate is $4 \times 10 ^ {-4}$. We employ dropout rate of 0.2 and batch normalization after graph information propgation. We also applied a dropout rate of 0.5 for RNNs and linear layers as regularization.  The hidden size of both story-line summarization RNN and decoder RNN is set to 512. The loss weight $\lambda_{gc}$ is set to 0.1.
At the inference stage, we use the greedy strategy to generate the final title.

%


\begin{figure}[!t] \begin{center}
    \includegraphics[width=\columnwidth]{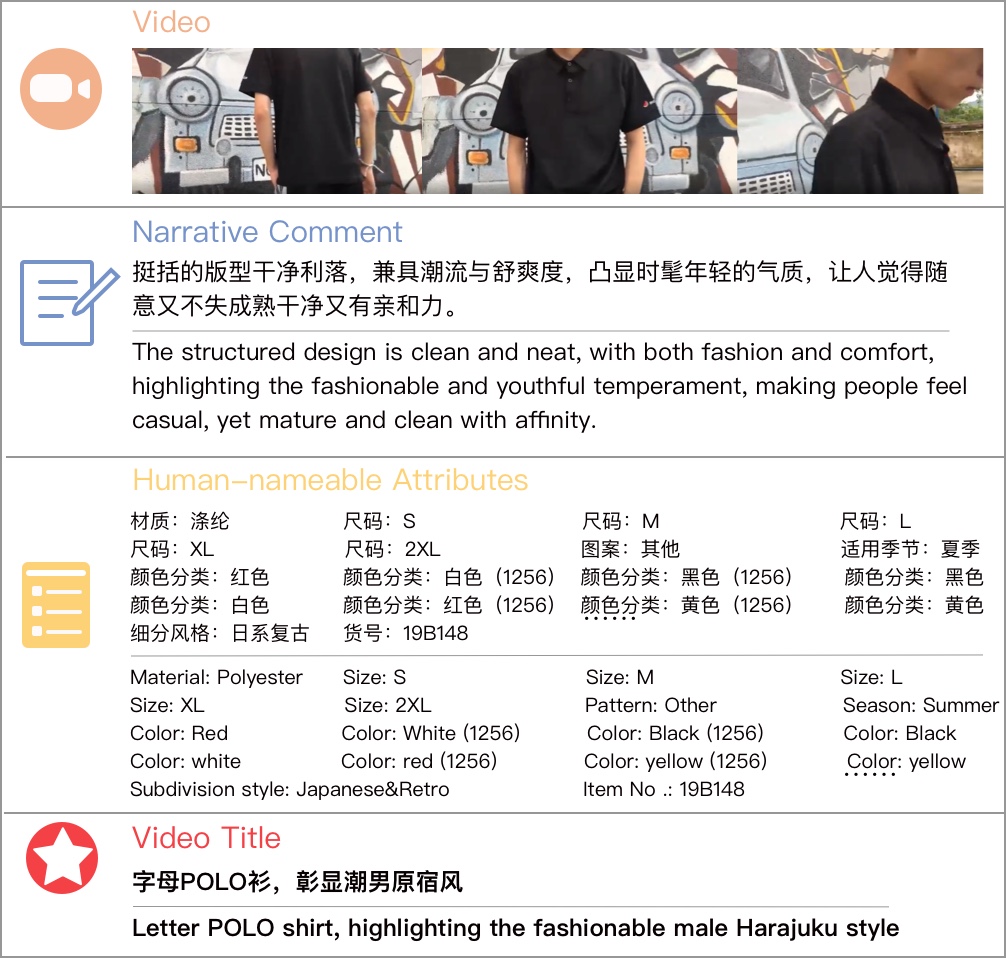}
    \caption{
    A <video, comment, attributes, title> quadruple data sample in T-VTD.
	}
\label{fig:example}
\end{center} \end{figure}

\vpara{Hardware \& Software Configuration }
The experiments are conducted on a Linux server equipped with an Intel(R) Xeon(R) CPU E5-2690 v4 @ 2.60GHz, 512GB RAM and 1 NVIDIA Titan V GPU.
All models are implemented in PyTorch \citep{paszke2017automatic} of version 1.2.0, pytorch-geometric \cite{Fey_Lenssen_2019} of version 1.3.2 and Python 3.6.
We'll soon release the main code of \Our for reproduction and further development.


\begin{table*}[t]
\centering
\setlength{\tabcolsep}{7.5pt}
\setlength\doublerulesep{0.5pt}
\begin{tabular}{l|cccccc}


\multicolumn{1}{c|}{Dataset} &    Context      &   \#Video    &   \#Sentence & \#Word & \#Vocabulary  &  Total Duration(hrs)   \\
\hline \hline

MSVD \cite{Chen_Dolan_2011}            & multi-category          &    1,970        &    70,028 & 607,339  & 13,010    & 5.3           \\

%

YouCook \cite{Das_Xu_Doell_Corso_2013}  & cooking     &  88  &  2,668  &   42,457  & 2,711  &  2.3 \\
TACos  \cite{Regneri_Rohrbach_Wetzel_Thater_Schiele_Pinkal_2013}   &  cooking    &  123  &  18,227  &  146,771  & 28,292 &  15.9
        \\
TACos M-L \cite{Rohrbach_Rohrbach_Qiu_Friedrich_Pinkal_Schiele_2014}  &  cooking  & 185   &  14,105  &  52,593  & - &  27.1
\\
MPII-MD \cite{Rohrbach_Rohrbach_Tandon_Schiele_2015}  &  movie      &  94  &  68,375  &  653,467   & 24,549 &  73.6
\\
M-VAD \cite{Torabi_Pal_Larochelle_Courville_2015}  &  movie        &  92  &  55,905   &  519,933   & 18,269  &  84.6
\\
VTW \cite{Torabi_Pal_Larochelle_Courville_2015}  &  multi-category        &  18,100  &  44,603   &  -   & 23,059  &  213.2
\\

MSR-VTT \cite{Zeng_Chen_Niebles_Sun_2016}  &  multi-category  & 7,180    & 200,000  &  1,856,523    & 29,316 &  41.2
\\
Charades \cite{Sigurdsson_Varol_Wang_Farhadi_Laptev_Gupta_2016}  &  human     &  9,848  & 27,847  & - & -  &  82.01
\\
\hline
\hline
T-VTD  &  e-commerce   & 90,000    & 180,000  & 3,878,436 & 68,232  &  755.73
\\
\hline

\end{tabular}
\caption{
Comparison between T-VTD with benchmark video captioning datasets, considering various capacity indicators.. 
}
\label{table:Data}
\end{table*}


\subsection{Additional Details on the Dataset } \label{sub:data}
The summarized statistics of T-VTD and comparisons with other frequently used benchmark video captioning datasets are shown in Figure \ref{table:Data}. Specifically, T-VTD contains 90000 videos with a total length of 755.73 hours, which is much larger than current datasets. We choose three best-sold categories, \ie, men wear, women wear and children wear, and collect 30,000 videos for each category. As for natural language data, T-VTD has totally 3,878,436 words (in titles and comments) with a vocabulary of 68232. The nature of abundant vocabulary and little repetitive information poses a direct challenge to fully understand the semantic information and avoid obtaining high scores just by overfitting to some biases. In contrast to existing datasets which mostly concern the object-level information (such as the object categories and actions), sentences in our dataset incorporate rich details in four levels: 1) landmark characteristics in the granular level; 2) the overall appearance and functionality in the product level; 3) the interaction with the background in the frame level 4) the story-line topic in the video level. These distinguishing features of T-VTD pose new challenges from the industry scenario. We summary the statistics of video durations in our dataset in Table \ref{table:videoduration}. The video durations in our dataset are mainly 15-30s, with an average video length around 30.23s. The longest video can reach 600 seconds (about 10 minutes). As shown in Table \ref{table:titlecontent} and \ref{table:attributes}, the average length of elements in the titles, comments and attributes are 6.6, 36.49 and 17.54, respectively. The vocabulary size is 27,976 for video titles and 59,109 for narrative comments. There are 81 different attribute-key types and 266,648 different attribute-value types, \ie, the vocabulary size of human-nameable attributes. A real case in our dataset is shown in Figure \ref{fig:example}. It can be seen that the comment sentences mainly narrate the preference for different aspects of products. Although the attributes of associated products structurally specify the human-nameable qualities of the products, these attributes can be noisy since they may cover all possible choices (such as different colors) and not exactly the product in the video. Models are required to fully understand the video rather than simply distill information from the attributes or the comment.



\begin{table}[!t]
\centering
\setlength{\tabcolsep}{5.5pt}
\setlength\doublerulesep{0.5pt}
\begin{tabular}{l|ccc}


  &    avg\_len      &   total\_len    &    vocab  \\
\hline \hline

title   &  6.6  & 594,279    & 27,976
\\
comment   &  36.49  & 3,284,157    & 59,109
\\

\hline

\end{tabular}
\caption{
    Basic statistics of titles and comments in T-VTD.
}
\label{table:titlecontent}
\end{table}


\subsection{Details on Data Pre-processing} \label{sub:id}

For text pre-processing, we remove the punctuations and tokenize sentences using Jieba Chinese Tokenizer \footnote{https://github.com/fxsjy/jieba}. Our vocabulary contains all attributes values, comment tokens and ground-truth title tokens. Since the real-world text data is noised and many expressions can be confusing or meaningless, such as brands and homophonic words. We roughly filter them by replacing low-frequency tokens (less than 50) with the special token $<unk>$, resulting in 6347 tokens in total. The length limitations for title, comment and attributes are 12, 50 and 15, respectively. Text with number of tokens more than the corresponding limitation will be truncated. We add a special $<sos>$ token as the first word for the title and a $<eos>$ at the end. When the $<sos>$ token is predicted in the decoding stage, the generation will be terminated.

For video processing, we first uniformly sample 30 frames per video. For landmark feature extraction, we extract the product area using internal product detector for all sampled frames. Then we use the pre-trained landmark detector \footnote{https://github.com/fdjingyuan/Deep-Fashion-Analysis-ECCV2018} provided by \cite{liu2018deep}. Specifically, the backbone model $VGG16$ takes each frame as input and output the activations of shape $512 \times 7 \times 7$ from layer \textit{pooled\_5}. This feature map is forwarded to the landmark decoder, which produce the landmark-oriented features of shape $64 \times 14 \times 14$ and the mask-like landmark maps of shape $8 \times 56 \times 56$, \ie, 8 landmarks maps of shape $\times 56 \times 56$ each. We downsampled each landmark map to have the same width and height as the landmark-oriented features. Deriving from the observation that intermediate feature map and emergent patterns are highly correlated, we normalize each landmark map as weights using softmax and compute weighted sum over the landmark-oriented features as the landmark feature.


\begin{table}[!t]
\centering
\setlength{\tabcolsep}{3.5pt}
\setlength\doublerulesep{0.5pt}
\begin{tabular}{l|cccccc}


  &    average  & min  & Q1   &   median   &  Q3 & max  \\
\hline \hline

video duration (s)  & 30.23  &  1.5   & 15.72  & 23.56    & 32.2  & 600.08
\\

\hline

\end{tabular}
\caption{
    Statistics of the video duration. (Q1 denotes the lower quartile and Q3 denotes the upper quartile.)
}
\label{table:videoduration}
\end{table}



For frame-level feature extraction, we use the same model as landmark feature extraction and obtain the activations of shape $128 \times 7 \times 7$ from layer \textit{conv4} for each frame. We use the global average pooling result as the frame feature.

The dataset we use for training is a subset (84394 samples) of the released dataset (90000 samples) due to the the data pre-processing (mainly the low-frequency words removal procedure) after which many words in the comment and elements in the attributes set will be replaced by $<unk>$. Specifically, we remove the following 3 kinds of samples: 1) sample with less than 2 non-$<unk>$ elements (\ie, where will be no edges in the graph) in the attributes set. 2) sample with 0 non-$<unk>$ words in the title. 3) sample with less than 11 non-$<unk>$ nodes in or less than 5 edges in the narrative comment graph. Overall, we mainly remove samples with little information within either one kind of fact (narrative comment or human-nameable attributes) or the ground truth title to make the data more reliable.

We randomly split the whole dataset by train 65\%, validation 5\% and test 30\%, resulting in 54856 samples for training, 4220 samples for validation and 25318 samples for testing.


\begin{table}[!t]
\centering
\setlength{\tabcolsep}{2.5pt}
\setlength\doublerulesep{0.5pt}
\begin{tabular}{l|ccccc}


  &    \multirow{2}{*}{avg\_num}      &   total\_num   &    total\_num  &  vocab & vocab  \\ & &  of keys & of values & of keys & of values  \\
\hline \hline

attributes   &  17.54  &  1,578,777    &  2,604,904  &  81 & 266,648
\\

\hline

\end{tabular}
\caption{
    Basic statistics of attributes in T-VTD.
}
\label{table:attributes}
\end{table}



